\documentclass[11pt, a4paper, logo, copyright, nonumbering]{deepseek}
\usepackage[numbers, square, sort&compress]{natbib}
\usepackage{dblfloatfix}
\usepackage{ulem}
\usepackage{caption}
\definecolor{citecolor}{HTML}{1976D2}
\hypersetup{
    colorlinks=true,
    linkcolor=black,
    filecolor=magenta,
    urlcolor=cyan,
    citecolor=citecolor,
}
\usepackage{dramatist}
\usepackage{xspace}
\usepackage{pifont}% http://ctan.org/pkg/pifont
\usepackage{multirow}
\usepackage{tcolorbox}
\usepackage{xltabular}
\usepackage{longtable}
\usepackage{hyperref}
\usepackage{diagbox}
\usepackage{makecell}
\usepackage{amsfonts}
\usepackage{amsmath}
\usepackage{amssymb}
\usepackage{lineno}

\usepackage[bottom]{footmisc}
\usepackage{CJKutf8}
\usepackage{subfigure}
\usepackage{setspace}
\usepackage{subcaption} 

\makeatletter
\def\@BTrule[#1]{%
  \ifx\longtable\undefined
    \let\@BTswitch\@BTnormal
  \else\ifx\hline\LT@hline
    \nobreak
    \let\@BTswitch\@BLTrule
  \else
     \let\@BTswitch\@BTnormal
  \fi\fi
  \global\@thisrulewidth=#1\relax
  \ifnum\@thisruleclass=\tw@\vskip\@aboverulesep\else
  \ifnum\@lastruleclass=\z@\vskip\@aboverulesep\else
  \ifnum\@lastruleclass=\@ne\vskip\doublerulesep\fi\fi\fi
  \@BTswitch}
\makeatother

\addto\extrasenglish{
}

 {\begin{list}{}%
         {\setlength{\leftmargin}{#1}}%
         \item[]%
 }
 {\end{list}}
 
\reportnumber{001} % Leave blank if n/a

\renewcommand{\today}{}

\title{\centering DeepSeek-OCR 2: Visual Causal Flow}

\author{Haoran Wei, Yaofeng Sun, Yukun Li\\
\small DeepSeek-AI\\
}
\renewcommand{\phi}{\varphi}

\renewcommand{\epsilon}{\varepsilon}
\renewcommand{\imath}{\mathrm{i}}

\newlength{\restsubwidth}
\newlength{\restsubheight}
\newlength{\restsubmoreheight}
\newcommand{\rest}[2]{%
        \settowidth{\restsubwidth}{\ensuremath{#2}}
        \settoheight{\restsubheight}{\ensuremath{{}_{#2}}}
        \ensuremath{{#1\hskip 0.5pt}_{\vrule\kern2pt\parbox[b][%
        4pt][b]{\the\restsubwidth}{%
                        \ensuremath{{}_{#2}}}}}
        }

\begin{abstract}
We present DeepSeek-OCR 2 to investigate the feasibility of a novel encoder—DeepEncoder V2—capable of dynamically reordering visual tokens upon image semantics. Conventional vision-language models (VLMs) invariably process visual tokens in a rigid raster-scan order (top-left to bottom-right) with fixed positional encoding when fed into LLMs. However, this contradicts human visual perception, which follows flexible yet semantically coherent scanning patterns driven by inherent logical structures. Particularly for images with complex layouts, human vision exhibits causally-informed sequential processing. Inspired by this cognitive mechanism, DeepEncoder V2 is designed to endow the encoder with causal reasoning capabilities, enabling it to intelligently reorder visual tokens prior to LLM-based content interpretation. This work explores a novel paradigm: whether 2D image understanding can be effectively achieved through two-cascaded 1D causal reasoning structures, thereby offering a new architectural approach with the potential to achieve genuine 2D reasoning. Codes and model weights are publicly accessible at \url{http://github.com/deepseek-ai/DeepSeek-OCR-2}.
\end{abstract}

\begin{document}
\begin{CJK*}{UTF8}{gbsn}

\maketitle

\begin{figure}[h]
    \centering
    \includegraphics[width=0.85\linewidth]{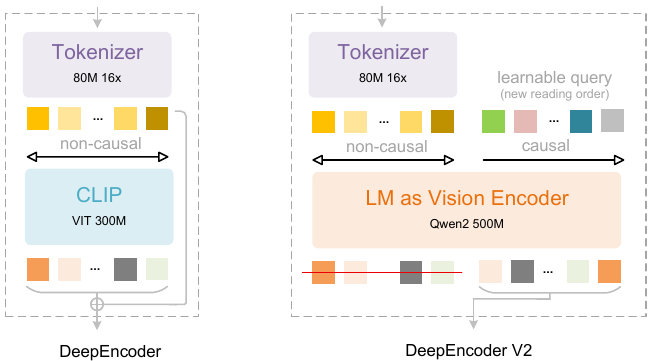}
    % \vspace{-0.2cm}
    \caption{We substitute the CLIP component in DeepEncoder with an LLM-style architecture. By customizing the attention mask, visual tokens utilize bidirectional attention while learnable queries adopt causal attention. Each query token can thus attend to all visual tokens and preceding queries, allowing progressive causal reordering over visual information.}
    \label{fig:1}
\end{figure}

\newpage

\begin{spacing}{0.9}
\tableofcontents
\end{spacing}

\newpage

\section{Introduction}

The human visual system closely mirrors transformer-based vision encoders~\cite{dosovitskiy2020image,dehghani2023patch}: foveal fixations function as visual tokens, locally sharp yet globally aware. However, unlike existing encoders that rigidly scan tokens from top-left to bottom-right, human vision follows a causally-driven flow guided by semantic understanding. Consider tracing a spiral—our eye movements follow inherent logic where each subsequent fixation causally depends on previous ones. By analogy, visual tokens in models should be selectively processed with ordering highly contingent on visual semantics rather than spatial coordinates.

This insight motivates us to fundamentally reconsider the architectural design of vision-language models (VLMs), particularly the encoder component. LLMs are inherently trained on 1D sequential data, while images are 2D structures. Directly flattening image patches in a predefined raster-scan order introduces unwarranted inductive bias that ignores semantic relationships. To address this, we present DeepSeek-OCR 2 with a novel encoder design—DeepEncoder V2—to advance toward more human-like visual encoding. Following DeepSeek-OCR~\cite{wei2025deepseek}, we adopt document reading as our primary experimental testbed. Documents present rich challenges including complex layout orders, intricate formulas, and tables. These structured elements inherently carry causal visual logic, demanding sophisticated reasoning capabilities that make document OCR an ideal platform for validating our approach.

Our main contributions are threefold:

First, we present DeepEncoder V2, featuring several key innovations: (1) we replace the CLIP~\cite{radford2021learning} component in DeepEncoder~\cite{wei2025deepseek} with a compact LLM~\cite{wang2024qwen2} architecture, as illustrated in Figure~\ref{fig:1}, to achieve visual causal flow; (2) to enable parallelized processing, we introduce learnable queries~\cite{carion2020end}, termed causal flow tokens, with visual tokens prepended as a prefix—through a customized attention mask, visual tokens maintain global receptive fields, while causal flow tokens can obtain visual token reordering ability; (3) we maintain equal cardinality between causal and visual tokens (with redundancy such as padding and borders) to provide sufficient capacity for re-fixation; (4) only the causal flow tokens—the latter half of the encoder outputs—are fed to the LLM~\cite{deepseekv2} decoder, enabling cascade causal-aware visual understanding.

Second, leveraging DeepEncoder V2, we present DeepSeek-OCR 2, which preserves the image compression ratio and decoding efficiency of DeepSeek-OCR while achieving substantial performance improvements. We constrain visual tokens fed to the LLM between 256 and 1120. The lower bound (256) corresponds to DeepSeek-OCR's tokenization of 1024×1024 images, while the upper bound (1120) matches Gemini-3 pro's~\cite{team2023gemini} maximum visual token budget. This design positions DeepSeek-OCR 2 as both a novel VLM architecture for research exploration and a practical tool for generating high-quality training data for LLM pretraining.

Finally, we provide preliminary validation for employing language model architectures as VLM encoders—a promising pathway toward unified omni-modal encoding. This framework enables feature extraction and token compression across diverse modalities (images, audio, text~\cite{liu2025context}) by simply configuring modality-specific learnable queries. Crucially, it naturally succeeds to advanced infrastructure optimizations from the LLM community, including Mixture-of-Experts (MoE) architectures, efficient attention mechanisms~\cite{deepseek32}, and so on.

In summary, we propose DeepEncoder V2 for DeepSeek-OCR 2, employing specialized attention mechanisms to effectively model the causal visual flow of document reading. Compared to the DeepSeek-OCR baseline, DeepSeek-OCR 2 achieves 3.73\% performance gains on OmniDocBench v1.5~\cite{ouyang2025omnidocbench} and yields considerable advances in visual reading logic.

\begin{figure}[ht]
	\centering
    \includegraphics[width=1.0\linewidth]{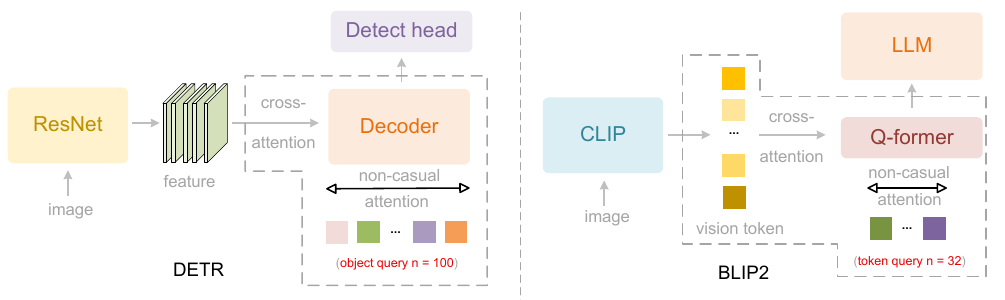}
    % \vspace{-2mm}
	\caption{This figure shows two computer vision models with parallelized queries: DETR's decoder~\cite{carion2020end} for object detection and BLIP2's Q-former~\cite{li2023blip} for visual token compression. Both employ bidirectional self-attention among queries.}
	\label{fig:encoders}
\end{figure}

\section{Related Works}

% \label{sec:related}

\subsection{Parallelized Queries in Decoder}

\label{sec:related1}

% Previously, object detection was dominated by two-stage detectors (e.g., Faster R-CNN~\cite{ren2015faster}) and one-stage detectors (e.g., YOLO~\cite{redmon2017yolo9000}), which heavily relied on hand-crafted components such as anchor boxes, non-maximum suppression (NMS), and feature pyramid networks (FPN~\cite{lin2017feature}).

DETR~\cite{carion2020end} pioneered the integration of transformer architecture into object detection, fundamentally breaking away from traditional detection paradigms~\cite{ren2015faster,redmon2017yolo9000}. To overcome the efficiency limitations of serial decoding in transformer blocks, DETR introduced preset parallelized learnable queries—a set of 100 object queries that encode object priors such as shape and position through training. These queries interact with feature maps~\cite{he2016deep} 
% extracted by the backbone network~\cite{he2016deep} 
via cross-attention mechanisms, while simultaneously engaging in bidirectional information exchange among themselves through self-attention. 
% The object queries output by the decoder are subsequently fed into detection heads for coordinate and category regression predictions. 
DETR established a foundational paradigm that enables transformers to handle parallelized tokens.
% , achieving truly end-to-end learning without the need for anchor design or NMS post-processing. 
The object query design has since become the de facto standard architectural component in subsequent transformer-based detection methods~\cite{zhu2020deformable,liu2021wb}.

\subsection{Parallelized Queries in Projector}
In recent years, vision-language models~\cite{li2023blip,Qwen-VL,Qwen2.5-VL,wei2024vary} have developed rapidly, with architectures converging toward the encoder-projector-LLM paradigm. The projector aligns visual tokens with the LLM's embedding space, serving as a critical bridge that enables LLMs to understand visual content. Q-former, introduced in BLIP-2~\cite{li2023blip}, exemplifies an effective projector design that employs learnable queries for visual token compression. Adopting a BERT-like~\cite{devlin2019bert} architecture and drawing inspiration from DETR's object queries~\cite{carion2020end}, Q-former utilizes 32 learnable queries that interact with hundreds of CLIP~\cite{radford2021learning} visual tokens through cross-attention. These compressed query representations are subsequently fed into the LLM, achieving effective mapping from visual to language space. The success of Q-former demonstrates that parallelized learnable queries are effective not only for feature decoding in detection tasks but also for token compression in multimodal alignment.

\subsection{LLM-based Multimodal Initialization}
LLMs trained on large-scale internet data have proven effective as initialization for multimodal models. Pang et al.~\cite{pang2023fozen} demonstrated that frozen LLM transformer layers enhance visual discriminative tasks. Moreover, encoder-free or lightweight-encoder models such as Fuyu~\cite{fuyu8b_model} and Chameleon~\cite{chameleon2024} in vision, as well as VALL-E~\cite{wang2023neural} in speech, further validate the potential of LLM pretrained weights for 
multimodal initialization.

\begin{figure}[ht]
	\centering
    \includegraphics[width=1.0\linewidth]{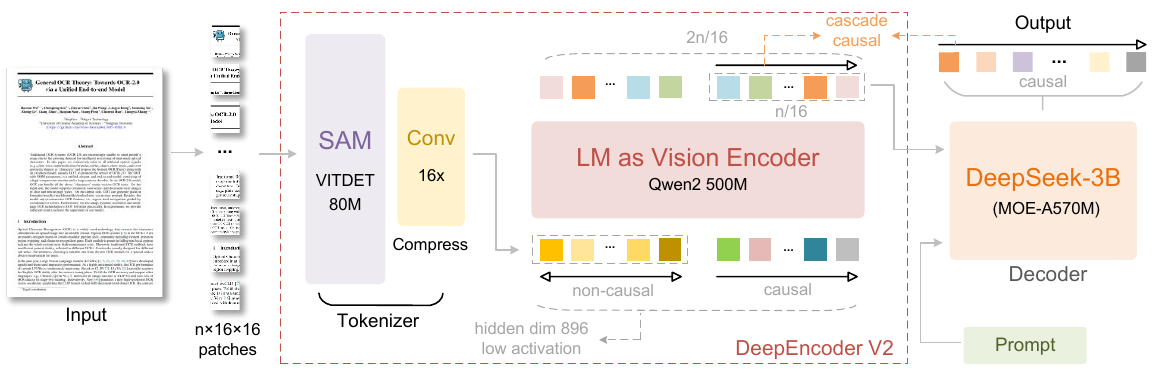}
    % \vspace{-2mm}
	\caption{DeepSeek-OCR 2 adopts the visual token compression mechanism from DeepEncoder, employing an 80M-parameter image compressor that reduces visual tokens by a factor of 16. DeepEncoder V2 differs by replacing DeepEncoder's CLIP module with a compact language model architecture. Through customized attention masks, this LM-style vision encoder acquires CLIP's knowledge compression capabilities while initiating causal modeling of visual sequences.}
	\label{fig:architecture}
\end{figure}

\section{Methodology}
\subsection{Architecture}
As shown in Figure~\ref{fig:architecture}, DeepSeek-OCR 2 inherits the overall architecture of DeepSeek-OCR, which consists of an encoder and a decoder. The encoder discretizes images into visual tokens, while the decoder generates outputs conditioned on these visual tokens and text prompts. The key distinction lies in the encoder: we upgrade DeepEncoder to DeepEncoder V2, which retains all capabilities of its predecessor while introducing causal reasoning through a novel architectural design. We elaborate on the details of DeepSeek-OCR 2 in the following sections.

\subsection{DeepEncoder V2}
The vanilla encoder serves as an important component that extracts and compresses image features through attention mechanisms, where each token attends to all others, achieving full-image receptive fields analogous to human foveal and peripheral vision. However, flattening 2D image patches into a 1D sequence imposes a rigid ordering bias through text-oriented positional encodings (e.g., RoPE~\cite{su2021roformer}). This contradicts natural visual reading patterns, especially non-linear layouts in optical texts, forms and tables.

\begin{figure}[ht]
	\centering
    \includegraphics[width=1.0\linewidth]{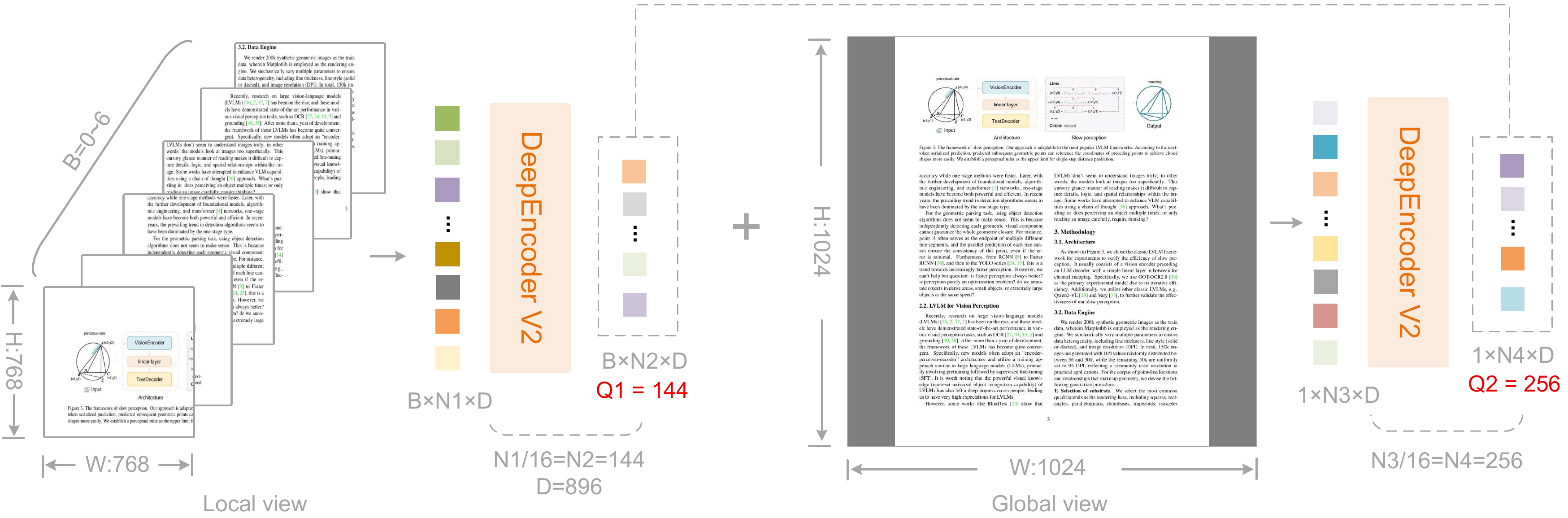}
    % \vspace{-2mm}
	\caption{Token count calculation in DeepEncoder V2. DeepEncoder V2 outputs 256$-$1120 tokens per image using a multi-crop strategy with 0$-$6 local views. With 0 local views, only the global view produces 256 tokens; with 6 local views, the count reaches 1120 (6$\times$144+256).}
	\label{fig:resolution}
\end{figure}

\subsubsection{Vision tokenizer}

The first component of DeepEncoder V2 is a vision tokenizer. Following DeepEncoder, we employ an architecture combining an 80M-parameter SAM-base~\cite{kirilloV2023segment} along with two convolutional layers~\cite{wei2024vary}. The output dimension of the final convolutional layer is reduced from 1024 in DeepEncoder to 896 to align with the subsequent pipeline. Note that this compression-based tokenizer is not mandatory and can be replaced with simple patch embedding. We retain it because it achieves 16$\times$ token compression~\cite{wei2024general,wang2025step,wei2024small,huang2026step3} through window attention with minimal parameters, significantly reducing both computational cost and activation memory for the subsequent global attention module. Moreover, its parameter count (~80M) remains comparable to the typical ~100M parameters used for text input embeddings in LLMs.

\subsubsection{Language model as vision encoder}
In DeepEncoder, a CLIP ViT follows the vision tokenizer to compress visual knowledge. DeepEncoder V2 redesigns this component into an LLM-style architecture with a dual-stream attention mechanism. Visual tokens utilize bidirectional attention to preserve CLIP's global modeling capability, while newly introduced causal flow queries employ causal attention. These learnable queries are appended after visual tokens as a suffix, where each query attends to all visual tokens and preceding queries. By maintaining equal cardinality between queries and visual tokens, this design imposes semantic ordering and distilling on visual features without altering token count. Finally, only the causal query outputs are fed to the LLM decoder.

We instantiate this architecture using Qwen2-0.5B~\cite{wang2024qwen2}, whose 500M parameters are comparable to CLIP ViT (300M) without introducing excessive computational overhead. The decoder-only architecture with prefix-concatenation of visual tokens proves crucial: extra experiments with cross-attention in an mBART-style~\cite{liu2020multilingual} encoder-decoder structure fail to converge. We hypothesize this failure stems from insufficient visual token interaction when isolated in a separate encoder. In contrast, the prefix design keeps visual tokens active throughout all layers, fostering effective visual information exchange with causal queries.

This architecture actually establishes two-stage cascade causal reasoning: the encoder semantically reorders visual tokens through learnable queries, while the LLM decoder performs autoregressive reasoning over the ordered sequence. Unlike vanilla encoders that impose rigid spatial ordering through positional encodings, our causally-ordered queries adapt to smooth visual semantics while naturally aligning with the LLM's unidirectional attention pattern. This design may bridge the gap between 2D spatial structure and 1D causal language modeling.

\begin{figure}[ht]
	\centering
    \includegraphics[width=0.8\linewidth]{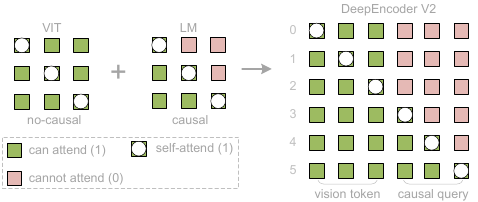}
    % \vspace{-2mm}
	\caption{Attention mask architecture of DeepEncoder V2. Concatenation of bidirectional mask (vision tokens, ViT-like) and causal triangular mask (flow tokens, LLM decoder-style).}
	\label{fig:attention}
\end{figure}

\subsubsection{Causal flow query}

As aforementioned, the number of causal query tokens equals the number of visual tokens, computed as $\frac{W \times H}{16^2 \times 16}$, where $W$ and $H$ denote the width and height of the image input to the encoder. To avoid maintaining multiple query sets for different resolutions, we adopt a multi-crop strategy with fixed query configurations at predefined resolutions.

Specifically, the global view uses a resolution of $1024 \times 1024$, corresponding to 256 query embeddings denoted as $\text{query}_{\text{global}}$. Local crops adopt a resolution of $768 \times 768$, with the number of crops $k$ ranging from 0 to 6 (no cropping is applied when both image dimensions are smaller than 768). All local views share a unified set of 144 query embeddings, denoted as $\text{query}_{\text{local}}$. Therefore, the total number of reordered visual tokens fed to the LLM is $k \times 144 + 256$, ranging from $[256, 1120]$. This maximum token count (1120) is lower than DeepSeek-OCR's 1156 (Gundam mode) and matches Gemini-3-Pro's maximum visual token budget.

\subsubsection{Attention mask}
To better illustrate the attention mechanism of DeepEncoder V2, we visualize the attention mask in Figure~\ref{fig:attention}. The attention mask is composed of two distinct regions. The left region applies bidirectional attention (similar to ViT) to original visual tokens, allowing full token-to-token visibility. The right region employs causal attention (triangular mask, identical to decoder-only LLMs) for causal flow tokens, where each token attends only to previous tokens. These two components are concatenated along the sequence dimension to construct DeepEncoder V2's attention mask (M), as follows:
\begin{equation}
M = \begin{bmatrix}
\mathbf{1}_{m \times m} & \mathbf{0}_{m \times n} \\
\mathbf{1}_{n \times m} & \text{LowerTri}(n)
\end{bmatrix},
\quad \text{where } n = m
\label{enq1}
\end{equation}
where $n$ is the number of causal query tokens, $m$ represents vanilla visual tokens number, and $\text{LowerTri}$ denotes a lower triangular matrix (with ones on and below the diagonal, zeros above).

\subsection{DeepSeek-MoE Decoder}

Since DeepSeek-OCR 2 primarily focuses on encoder improvements, we do not upgrade the decoder component. Following this design principle,  we retain DeepSeek-OCR's decoder $-$ a 3B-parameter MoE structure with about 500M active parameters. The core forward pass of DeepSeek-OCR 2 can be formulated as:
\begin{equation}
\mathbf{O} = \mathcal{D}\left(\pi_Q\left(\mathcal{T}^L\left(\mathcal{E}(\mathbf{I}) \oplus \mathbf{Q}_0; \mathbf{M}\right)\right)\right)
\label{enq2}
\end{equation}
where $\mathbf{I} \in \mathbb{R}^{H \times W \times 3}$ is the input image, $\mathcal{E}$ is the vision tokenizer mapping images to $m$ visual tokens $\mathbf{V} \in \mathbb{R}^{m \times d}$, $\mathbf{Q}_0 \in \mathbb{R}^{n \times d}$ are learnable causal query embeddings, $\oplus$ denotes sequence concatenation, $\mathcal{T}^L$ represents an $L$-layer Transformer with masked attention, $\mathbf{M} \in \{0,1\}^{2n \times 2n}$ is the block causal attention mask defined in Equation~\ref{enq1}, $\pi_Q$ is the projection operator that extracts the last $n$ tokens (i.e., $\mathbf{Z} = \mathbf{X}_{m+1:m+n}$), $\mathcal{D}$ is the language decoder, and $\mathbf{O} \in \mathbb{R}^{n \times |\mathcal{V}|}$ is the output logits over LLM vocabulary.

\section{Experimental Settings}

\subsection{Data Engine}
\label{data}

DeepSeek-OCR 2 employs the same data sources as DeepSeek-OCR, comprising OCR 1.0, OCR 2.0~\cite{chen2024onechart,wei2024slow,liu2024focus_fox}, and general vision data~\cite{wei2025deepseek}, with OCR data constituting 80\% of the training mixture. We also introduce two modifications: (1) a more balanced sampling strategy for OCR 1.0 data, partitioning pages by content type (text, formulas, tables) with a 3:1:1 ratio, and (2) label refinement for layout detection by merging semantically similar categories (e.g., unifying "figure caption" and "figure title"). Given these minimal differences, we consider DeepSeek-OCR a valid baseline for comparison.

\subsection{Training Pipelines}
We train DeepSeek-OCR 2 in three stages: (1) encoder pretraining, (2) query enhancement, and (3) decoder specialization. The stage-1 enables the vision tokenizer and LLM-style encoder to acquire fundamental capabilities in feature extraction, token compression, and token reordering capabilities. The stage-2 further strengthens the token reordering capability of the encoder while enhancing visual knowledge compression. The stage-3 freezes the encoder parameters and optimizes only the decoder, enabling higher data throughput under the same FLOPs.

\subsubsection{Training DeepEncoder V2}
Following DeepSeek-OCR and Vary~\cite{wei2024vary}, we train DeepEncoder V2 using a language modeling objective, coupling the encoder with a lightweight decoder~\cite{OPT-IML} for joint optimization via next token prediction. We employ two dataloaders at resolutions of 768$\times$768 and 1024$\times$1024. The vision tokenizer is initialized from DeepEncoder, and the LLM-like encoder from Qwen2-0.5B-base~\cite{wang2024qwen2}. After pretraining, only the encoder parameters are retained for subsequent stages. We use the AdamW~\cite{AdamW} optimizer with cosine learning rate decay from 1e-4 to 1e-6, training on 160 A100 GPUs (20 nodes $\times$ 8 GPUs) with batch size 640 for 40k iterations (with sequence packing at 8K length, about 100M image-text pair samples).

\subsubsection{Query enhancement}
After DeepEncoder V2 pretraining, we integrate it with DeepSeek-3B-A500M~\cite{deepseekv2,deepseekv3} as our final pipeline. We freeze the visual tokenizer (SAM-conv structure) while jointly optimizing the LLM encoder and LLM decoder to enhance query representations. At this stage, we unify the two resolutions into a single dataloader via multi-crop strategy. We adopt 4-stage pipeline parallelism: vision tokenizer (PP0), LLM-style encoder (PP1), and DeepSeek-LLM layers (6 layers per stage on PP2-3). With 160 GPUs (40GB/per-GPU), we configure 40 data parallel replicas (4 GPUs per replica) and train with global batch size 1280 using the same optimizer and learning rate decay from 5e-5 to 1e-6 over 15k iterations.

\subsubsection{Continue-training LLM}
To rapidly consume training data, we freeze all DeepEncoder V2 parameters in this stage and only update the DeepSeek-LLM parameters. This stage accelerates training (more than doubles the training speed under the same global batch size) while helping the LLM better understand DeepEncoder V2's reordered visual tokens. Continuing from stage-2, we perform another learning rate decay from 1e-6 to 5e-8 training for 20k iterations in this stage.

\begin{table}[h]
\footnotesize
\caption{Comprehensive evaluation of document reading on OmniDocBench v1.5. V-token$^{max}$ represents the maximum number of visual tokens used per page in this benchmark. R-order denotes reading order. Except for DeepSeek OCR and DeepSeek OCR 2, all other model results in this table are sourced from the OmniDocBench repository. }
\label{tab:omnidocbench}
\centering
%\resizebox{1.\linewidth}{!}
\setlength{\tabcolsep}{1.6pt}
{
\begin{tabular}{l|c|c|ccccc}
\toprule[.9pt]
{\textbf{Model}} & V-token$^{max}$ $\downarrow$ & Overall $\uparrow$ & Text$^{Edit}$ $\downarrow$  &  Formula$^{CDM}$ $\uparrow$ & Table$^{TEDs}$ $\uparrow$ & Table$^{TEDS_s}$ $\uparrow$ & R-order$^{Edit}$ $\uparrow$  \\  
\midrule  
% \shline
\multicolumn{8}{c}{\textbf{  Pipline}} \\ 
\midrule 
Marker-1.8.2~\cite{marker} & - & 71.30 & 0.206 & 76.66 & 57.88 & 71.17 & 0.250 \\
MinerU2-pp~\cite{wang2024mineru} & - & 71.51 & 0.209 & 76.55 & 70.90  & 79.11 & 0.225\\ 
Dolphin~\cite{feng2025dolphin} & - & 74.67 & 0.125 & 67.85 & 68.70  & 77.77 & 0.124\\ 
Dolphin-1.5~\cite{feng2025dolphin} & - & 83.21 & 0.092 & 80.78 & 78.06  & 84.10 & 0.080\\ 
PP-StructureV3~\cite{cui2025paddleocr} & - & 86.73 & 0.073 & 85.79 & 81.68 & 89.48 & 0.073 \\ 
MonkeyOCR-pro-1.2B~\cite{li2025monkeyocr} & - & 86.96 & 0.084 & 85.02 & 84.24 & 89.02 & 0.130 \\ 
MonkeyOCR-3B~\cite{li2025monkeyocr} & - & 87.13 & 0.075 & 87.45 & 81.39 & 85.92 & 0.129 \\ 
MonkeyOCR-pro-3B~\cite{li2025monkeyocr} & - & 88.85 & 0.075 & 87.25 & 86.78 & 90.63 & 0.128 \\ 
MinerU2.5~\cite{wang2024mineru} & - & 90.67 & 0.047 & 88.46 & 88.22 & 92.38 & 0.044 \\ 
PaddleOCR-VL~\cite{cui2025paddleocrvl} & - & 92.86 & 0.035 & 91.22 & 90.89 & 94.76 & 0.043 \\ 
\midrule 
\multicolumn{8}{c}{\textbf{    End-to-end Model}} \\ 
\midrule 
OCRFlux~\cite{ocrflux} & >6000 & 74.82 & 0.193 & 68.03 & 75.75 & 80.23 & 0.202 \\
GPT-4o~\cite{GPT4} & - & 75.02 & 0.217 & 79.70 & 67.07 & 76.09 & 0.148 \\
InternVL3~\cite{zhu2025internvl3}& >7000 & 80.33 & 0.131 & 83.42 & 70.64 & 77.74  & 0.113 \\
POINTS-Reader~\cite{liu2025pointsreader} & >6000 & 80.98 & 0.134 & 79.20 & 77.13 & 81.66 & 0.145 \\
olmOCR~\cite{poznanski2025olmocr} & >6000 & 81.79 & 0.096 & 86.04 & 68.92 & 74.77 & 0.121 \\
InternVL3.5-241B~\cite{wang2025internvl35}& >7000 & 82.67 & 0.142 & 87.23 & 75.00 & 81.28  & 0.125 \\
MinerU2-VLM~\cite{wang2024mineru}& >7000 & 85.56 & 0.078 & 80.95 & 83.54 & 87.66  & 0.086 \\
Nanonets-OCR-s~\cite{NanonetsOCRs}& >7000 & 85.59 & 0.093 & 85.90 & 80.14 & 85.57  & 0.108 \\
Qwen2.5-VL-72B~\cite{Qwen2.5-VL} & >6000 & 87.02 & 0.094 & 88.27 & 82.15 & 86.22 & 0.102 \\
Gemini-2.5 Pro\cite{google_gemini_web} & - & 88.03 & 0.075 & 85.82 & 85.71 & 90.29 & 0.097 \\
dots.ocr~\cite{dots} & >6000 & 88.41 & 0.048 & 83.22 & 86.78 & 90.62 & 0.053 \\ 
OCRVerse~\cite{OCRVerse} & >6000 & 88.56 & 0.058 & 86.91 & 84.55 & 88.45 & 0.071 \\ 
Qwen3-VL-235B~\cite{bai2025qwen3vltechnicalreport} & >6000 & 89.15 & 0.069 & 88.14 & 86.21 & 90.55 & 0.068 \\ 
\midrule 
% \rowcolor{gray!10}
DeepSeek-OCR (9-crops) & 1156 & 87.36 & 0.073 & 84.14 & 85.25 & 89.01 & 0.085 \\ 
% \rowcolor{gray!10}
DeepSeek-OCR 2 & 1120 & 91.09  & 0.048  & 90.31  & 87.75  & 92.06 &  0.057 \\ 

 &  \textcolor{blue}{$\downarrow 36$} &\textcolor{blue}{$\uparrow 3.73$} & \textcolor{blue}{$\downarrow 0.025$} & \textcolor{blue}{$\uparrow 6.17$} & \textcolor{blue}{$\uparrow 2.5$} & \textcolor{blue}{$\uparrow 3.05$} &\textcolor{blue}{$\downarrow 0.028$}\\
\bottomrule[.9pt]
\end{tabular}
% \vspace{-.6em}
}
% \vspace{-2.em}
\end{table}

\section{Evaluation}

We select OmniDocBench v1.5~\cite{ouyang2025omnidocbench} as our primary benchmark for evaluation. This benchmark comprises 1,355 document pages spanning 9 major categories (including magazines, academic papers, research reports, and so on) in both Chinese and English. With its diverse test samples and robust evaluation criteria, OmniDocBench provides an effective framework for validating the performance of DeepSeek-OCR 2, particularly the effectiveness of DeepEncoder V2.

\subsection{Main Results}
\label{OmniDocbench v1.5}

As shown in Table~\ref{tab:omnidocbench}, DeepSeek-OCR 2 achieves advanced performance of 91.09\% while using the smallest upper limit of visual tokens (V-token$^{max}$). Compared to the DeepSeek-OCR baseline, it demonstrates a 3.73\% improvement under similar train data sources, validating the effectiveness of our newly designed architecture. Beyond the overall improvement, the Edit Distance (ED) for reading order (R-order) has also significantly decreased (from 0.085 to 0.057), indicating that the new DeepEncoder V2 can effectively select and arrange initial visual tokens based on image information. As illustrated in Table~\ref{table2}, DeepSeek-OCR 2 (0.100) achieves lower ED in document parsing compared to Gemini-3 Pro (0.115) under a similar visual token budget (1120), further demonstrating that our new model maintains high compression rates of visual tokens while ensuring superior performance, with exceptionally high potential.

\begin{table}[!t]\small
	\centering	
        
 	\caption{Edit Distances for different categories of document-elements in OmniDocBench v1.5. V-token$^{max}$ denotes the lowest maximum number of visual tokens.}
    \setlength{\abovecaptionskip}{0.2cm}
	\setlength{\tabcolsep}{1.5mm}{	
		
            \begin{tabular}{l|c|cccc|c}  %
            \toprule 
            Model & V-token$^{max}$$\downarrow$ & Text$^{Edit}$$\downarrow$ & Formula$^{Edit}$$\downarrow$ & Table$^{Edit}$$\downarrow$ & R-order$^{Edit}$$\downarrow$ & Overall$^{Edit}$$\downarrow$ \\ 
            \midrule
			Gemini-3 pro~\cite{team2023gemini}& 1120 & - & -  & - & - & 0.115  \\
            Seed-1.8~\cite{seedseed1}& 5120 & - & -  & - & - & 0.106  \\
            DeepSeek-OCR& 1156 & 0.073 & 0.236  & 0.123 & 0.085 & 0.129  \\
            DeepSeek-OCR 2& 1120 &0.048 & 0.198  & 0.096  & 0.057  & 0.100   \\
            % Process& resize & resize &  & \\

			\bottomrule		
	\end{tabular}}		

	\label{table2}
\end{table}

\subsection{Improvement Headroom}

We conduct a detailed performance comparison between DeepSeek-OCR and DeepSeek-OCR 2 across 9 document types and found that DeepSeek-OCR 2 still has considerable room for improvement, as shown in Table~\ref{table-3}. For text recognition Edit Distance (ED), DeepSeek-OCR 2 outperforms DeepSeek-OCR in most cases, but there are also notable weaknesses, such as newspapers, where it performs $>0.13$ ED. We believe there are two main reasons: (1) the lower upper limit of visual tokens may affect the recognition of text-super-rich newspapers, which can be simply addressed in the future by increasing the number of local crops; (2) insufficient newspaper data $-$ our training data contains only 250k relevant samples, which is inadequate for training DeepEncoder V2 for this class. Of course, for the reading order (R-order) metric, DeepSeek-OCR 2 consistently outperforms DeepSeek-OCR across the board, further validating the effectiveness of our visual causal flow encoder design.

\begin{table}[ht]\small
    \centering	
	% \begin{center}

    \caption{Detailed comparison between DeepSeek-OCR 2 and DeepSeek-OCR across 9 document types. R-order denotes reading order. All metrics are Edit Distances, where lower is better.}
    % \resizebox{7.5cm}{!}{
    \setlength{\abovecaptionskip}{0.2cm}
    \setlength{\tabcolsep}{0.6mm}{

        \begin{tabular}{l|c|ccccccccc}
            % \hline  
                % \toprule
            % \hline
                \toprule
            %	\rowcolor{gray!20}
            Model & Edit $\downarrow$ & PPT & \makecell{Academic \\ Paper} & Book & \makecell{Colorful \\ Textbook} & \makecell{Exam \\ Paper} & Magazine & Newspaper & Note & \makecell{Research \\ Report}  \\
            % \hline  % 中部线
                \midrule
            % \rowcolor{gray!10}
                \multirow{2}{*}{DS-OCR} & Text & 0.052 & 0.028 & 0.022 & 0.130 & 0.074 & 0.049 & 0.131 & 0.145 & 0.015   \\
                % \cmidrule{2-2}
                  & R-order & 0.052 & 0.021 & 0.040 & 0.125 & 0.083 & 0.101 & 0.217 & 0.089 & 0.016   \\
                % \cmidrule{1-6}
                \midrule
                 \multirow{2}{*}{DS-OCR 2}  &Text & 0.031 & 0.013 & 0.033 & 0.053 & 0.047 & 0.026 & 0.139 & 0.068 & 0.008  \\
                 % \cmidrule{2-11}
                         &R-order & 0.025 & 0.013 & 0.027 & 0.066 & 0.048 & 0.100 & 0.176 & 0.035 & 0.011   \\

    % \hline
    \bottomrule
    \end{tabular}}

	% }	
	% \end{center}
    \label{table-3}
	
\end{table}

\begin{table}[!t]\small
    \centering

    \caption{Production performance comparison between DeepSeek-OCR and DeepSeek-OCR 2. For OCR models serving LLM pipelines, ground truth is not accessible in production environments. Therefore, Repetition rate constitutes the primary observable quality metric.}
    % \resizebox{7.5cm}{!}{
    \setlength{\abovecaptionskip}{0.2cm}
    \setlength{\tabcolsep}{1.5mm}{

        \begin{tabular}{l|c|cc}
            % \hline  
                % \toprule
            % \hline
                \toprule
            %	\rowcolor{gray!20}
            Model & Metric & online-user-logs (image) & pretrain-data (PDF)   \\
            % \hline  % 中部线
                \midrule
            % \rowcolor{gray!10}
                \multirow{1}{*}{DeepSeek-OCR} & \multirow{2}{*}{Repeat $\downarrow$} & 6.25\% & 3.69\%    \\
                % \cmidrule{2-2}
                
                \cmidrule{1-1}
                \cmidrule{3-4}
                % \midrule
                 \multirow{1}{*}{DeepSeek-OCR 2}  & &4.17\% \textcolor{blue}{$\downarrow 2.08\%$}  & 2.88\% \textcolor{blue}{$\downarrow 0.81\%$}  \\

    % \hline
    \bottomrule
    \end{tabular}}

	% }	

	\label{table4}
    \end{table}

\subsection{Practical Readiness }

DeepSeek-OCR serves two primary production use cases: an online OCR service that reads image/documents for DeepSeek-LLMs, and a pretraining data pipeline that performs batch PDF processing. We compare the production performance between DeepSeek-OCR 2 and DeepSeek-OCR. Since ground truth is unavailable in production environments, we focus primarily on repetition rate as our key metric. As shown in Table~\ref{table4}, DeepSeek-OCR 2 demonstrates markedly  improved practical readiness compared to its predecessor (DeepSeek-OCR), reducing the repetition rate from 6.25\% to 4.17\% for online user-log images, and from 3.69\% to 2.88\% for PDF data production. These results further validate the effectiveness of the DeepSeek-OCR 2 architecture, particularly its logical visual comprehension capabilities.

\section{Discussion and Future Works}
\subsection{Towards Genuine 2D Reasoning}
DeepSeek-OCR 2 presents a novel architectural paradigm with an LLM-style encoder cascaded with an LLM decoder. This cascade of two 1D causal reasoners holds promise for genuine 2D reasoning: the encoder performs reading logic reasoning (causally reordering visual information through query tokens), while the decoder executes visual task reasoning over these causally-ordered representations. Decomposing 2D understanding into two complementary/orthogonal 1D causal reasoning subtasks may represent a breakthrough toward genuine 2D reasoning. Of course, achieving this goal remains a long journey. For instance, to enable multiple re-examinations and multi-hop reordering  of visual content, we may need substantially longer causal flow tokens than the original visual token sequence. We will continue to refine this architecture and explore its effectiveness on general visual reasoning tasks in future work.

\subsection{Towards Native Multimodality}

DeepEncoder V2 provides initial validation of the LLM-style encoder's viability for visual tasks. More importantly, this architecture enjoys the potential to evolve into a unified omni-modal encoder: a single encoder with shared $Wk, Wv$ projections, attention mechanisms, and FFNs can process multiple modalities through modality-specific learnable query embeddings. Such an encoder could compress text, extract speech features, and reorganize visual content within the same parameter space, differing only in the learned weights of their query embeddings. DeepSeek-OCR's optical compression represents an initial exploration toward native multimodality, while we believe DeepSeek-OCR 2's LLM-style encoder architecture marks our further step in this direction. We will also continue exploring the integration of additional modalities through this shared encoder framework in the future.

\section{Conclusion}
In this technical report, we present DeepSeek-OCR 2, a significant upgrade to DeepSeek-OCR, that maintains high visual token compression while achieving meaningfully performance improvements. This advancement is powered by the newly proposed DeepEncoder V2, which implicitly distills causal understanding of the visual world through the integration of both bidirectional and causal attention mechanisms, leading to causal reasoning capabilities in the vision encoder and, consequently, marked lifts in visual reading logic.

While optical text reading, particularly document parsing, represents one of the most practical vision tasks in the LLM era, it constitutes only a small part of the broader visual understanding landscape. Looking ahead, we will refine and adapt this architecture to more diverse scenarios, seeking deeper toward a more comprehensive vision of multimodal intelligence.

\newpage

\bibliography{main}

\end{CJK*}
\end{document}